\documentclass[nohyperref]{article}

\usepackage{microtype}
\usepackage{graphicx}
\usepackage{subfigure}
\usepackage{booktabs}    
\usepackage{multirow}
\usepackage{array}

\usepackage{hyperref}
\usepackage{url}

\usepackage[accepted]{icml2022}

\makeatletter
\renewcommand{\ICML@appearing}{\textit{Proceedings of the
Second Workshop on Technical AI Governance Research (TAIGR)
at the $\mathit{43}^{rd}$ International Conference on Machine Learning},
Seoul, South Korea, 2026.
Copyright 2026 by the author(s).}

\renewcommand{\printAffiliationsAndNotice}[1]{%
\stepcounter{@affiliationcounter}%
{\let\thefootnote\relax\footnotetext{\hspace*{-\footnotesep}%
\forloop{@affilnum}{1}{\value{@affilnum} < \value{@affiliationcounter}}{
\textsuperscript{\arabic{@affilnum}}\ifcsname @affilname\the@affilnum\endcsname%
\csname @affilname\the@affilnum\endcsname%
\else
{\bf AUTHORERR: Missing \textbackslash{}icmlaffiliation.}
\fi
}.
\ifdefined\isaccepted #1\fi%
\ifdefined\icmlcorrespondingauthor@text
Correspondence to: \icmlcorrespondingauthor@text.
\else
{\bf AUTHORERR: Missing \textbackslash{}icmlcorrespondingauthor.}
\fi

\ \\
\Notice@String
}
}
}
\makeatother

\usepackage{amsmath}
\usepackage{amssymb}
\usepackage{mathtools}
\usepackage{amsthm}

\usepackage[capitalize,noabbrev]{cleveref}

\icmltitlerunning{Detecting Hidden ML Training With Zero-Overhead Telemetry}

\begin{document}

\twocolumn[
\icmltitle{Detecting Hidden ML Training With \\
           Zero-Overhead Telemetry}

\icmlsetsymbol{equal}{*}

\begin{icmlauthorlist}
\icmlauthor{Robi Rahman}{miri}
\icmlauthor{Sabiha Tajdari}{uva}
\end{icmlauthorlist}

\icmlaffiliation{miri}{Machine Intelligence Research Institute}
\icmlaffiliation{uva}{University of Virginia}

\icmlcorrespondingauthor{Robi Rahman}{robi@intelligence.org}

\icmlkeywords{compute governance, AI governance, GPU telemetry, workload
classification, adversarial robustness, hardware-enabled mechanisms}

\vskip 0.3in
]

\printAffiliationsAndNotice{Equal contribution. }

\begin{abstract}
Hardware-enabled monitoring of GPU workloads underpins many proposals for AI
compute governance, but if developers can defeat monitoring mechanisms, such
schemes are unworkable. We evaluate the adversarial robustness of GPU workload
classification using only zero-overhead, privacy-preserving NVML telemetry:
content-agnostic signals that observe physical effects of computation without
accessing model weights, training data, or hyperparameters. Across 5 rounds of
monitor--evader iteration, we evaluate 20 evasion strategy families on 9 GPU
models spanning 4 architecture generations. We develop a classifier that
achieves 98.2\% binary accuracy at identifying training workloads across the
whole corpus, and 43--87\% accuracy against the most challenging unexpected
workloads even when they are adversarially disguised.
\end{abstract}

\section{Introduction}
\label{sec:intro}

Compute governance proposals such as \citet{shavit2023}, \citet{baker2025}, and
\citet{scher2025} call for regulations including evaluations, audits, and
registration or prohibition of training runs above specified compute thresholds
\citep{heim2024thresholds}. U.S.\ Executive Order 14110, in effect from October
2023 until rescinded in January 2025, previously required developers which
perform training runs exceeding $10^{26}$ floating-point operations (FLOP) to
report those runs to the federal government, along with their cybersecurity and
risk management practices \citep{eo14110}.

However, the ability to regulate ML training compute requires that regulators
are able to know when training occurs, and that developers cannot discreetly run
training workloads undetected. Work is underway to establish the feasibility of
geolocation of computing hardware \citep{brass2024} and registries of AI chips
\citep{fist2023}. But registration and geolocation are insufficient unless the
regulator understands what the chips are doing.

Past work demonstrates the feasibility of distinguishing ML from non-ML
computing when the adversary is not trying to disguise the workloads
\citep{tang2022, weakley2025}, and existing governance literature discusses
masking or structuring as possible evasion
tactics \citep{heim2024cloud, seferis2025}. However, there has not yet
been a systematic effort to iterate on this cat-and-mouse dynamic and determine
whether the problem landscape favors detection or evasion.

Therefore, we investigate the adversarial robustness of GPU workload
classifiers, trained on signals extracted from built-in telemetry readings. We
build the classifier as a prerequisite, then extensively red-team it with 20+
categories of real and adversarial workloads across 100+ GPU-hours of ML
experiments.

We find that zero-overhead GPU telemetry is sufficient to classify workloads as
training vs.\ non-training with 98.2\% accuracy across 9 GPU types spanning
consumer and datacenter hardware (Ampere through Blackwell). Crucially, this
telemetry is content-agnostic and privacy-preserving: it observes physical
effects of computation (power, temperature, utilization) without accessing model
weights, training data, or hyperparameters \citep{shavit2023, ogara2025}.

Through 5 rounds of monitor--evader iteration, we implement 20 families of
evasion strategies, including 4 multi-GPU model- and/or data-parallel
configurations on nodes of up to 8 datacenter cards. Leave-one-strategy-out
evaluation---training classifiers on some evasion tactics while evaluating on
unknown workloads---shows mean accuracy of 95.2\% on unseen external strategies
(92.2\% including classifier-assisted white-box attacks), with all multi-GPU
workloads detected at $\geq$99.6\% and only 5 of 20 families under 90\%
detection against the best classifier. Physics-based evasion-resistant features
(cumulative energy, power--memory correlation, pre-allocation memory-plateau
features) provide the signal needed to catch the hardest remaining single-GPU
strategies.

We release a privacy-preserving, content-agnostic classifier achieving 98.2\%
accuracy on training, inference, and non-ML workloads run on NVIDIA hardware in
generations Ampere through Blackwell, using only zero-overhead NVIDIA Management
Library (NVML) telemetry signals. We evaluate cross-architecture generalization,
demonstrating that classifiers trained on a representative variety of hardware
detect training on new, unseen hardware without re-fitting. Source code for the
non-ML, inference and training, and adversarial evasion workloads is available
in the project repository. We invite the AI governance and ML research
communities to red-team our classifier and publish new evasion strategies that
might defeat it, which will make it even more robust in the future.

\section{Background and Threat Model}
\label{sec:background}

\subsection{Telemetry Signals}
\label{sec:telemetry}

Our classifier operates on nine always-available NVML counters (GPU utilization,
memory utilization, memory used, power draw, temperature, SM clock, memory
clock, PCIe TX/RX) sampled at 1\,Hz. These counters are available on every
NVIDIA GPU from Pascal onwards with no runtime overhead: polling them does not
measurably burden or slow down the monitored workload, and does not require any
additional configuration. The readings do not involve access to application
code, model weights, or training data---these signals are purely physical or
electrical measurements of utilization of different components of the GPU.

By contrast, existing profiling-based monitoring proposals carry prohibitive
costs. WAVE \citep{xu2026} achieves privacy-preserving oversight using Nsight
Compute to read performance counters, but this adds 1200--5300\% runtime
overhead: Nsight Compute is a kernel-debugging profiler not designed for
production monitoring, and is very inefficient to run continuously along with
workloads. NVML counters are a more efficient choice, and more sustainable since
datacenters are already using huge and rapidly increasing amounts of
energy \citep{epri2024, chen2025grid}. Our result establishes a lower bound on
the efficacy of convenient, economical oversight: anything that works with this
minimal observational budget also works with more information, so a real
deployment monitor that can access more telemetry or physical channels starts
from this demonstrated detection level and only becomes stronger.

\subsection{Threat Model}
\label{sec:threat}

We consider two types of evaders. One is a \emph{cloud tenant}: a user or AI
developer renting GPU time, attempting to disguise ML training as inference or
non-ML workloads to evade the cloud provider's monitoring. Proposals such as
\citet{heim2024cloud} rely on the intermediary role of compute providers acting
as ``record keepers, verifiers, and, in some cases, even enforcers'' of AI
governance. In this case, the adversary controls user-space software but not the
host OS, hypervisor, or hardware. One type of evasion they might attempt is
\emph{compute structuring}, wherein training runs are split across multiple
cloud providers to dodge a reporting threshold. \citet{seferis2025} argue that
these attacks are detectable, but this relies on the ability of each cloud host
to reliably classify the workloads running on its own GPUs. If tenants can
disguise training as other workloads and evade recognition by the host, the
compute-structuring defense collapses. Our classifier provides the foundation
that allows this type of enforcement to work, and our adversarial evaluation is
a test of its robustness.

Another type of evader is a \emph{hardware operator}. They own or operate the
hardware, typically in a datacenter facility, and attempt to hide unauthorized
AI training from an external regulator. The adversary controls the full software
stack and can modify workload scheduling. In order to prevent this evader from
defeating the monitoring scheme, we need to equip their chips with a
\emph{hardware-enabled governance mechanism} (HEM) that ensures that the evader
cannot (i) disable NVML polling or tamper with the returned counter values at
the hardware level, nor (ii) inject synthetic telemetry into the monitor's
ingestion channel. Both assumptions describe mechanisms the software stack alone
cannot enforce; they are the minimum hardware-enabled-mechanism (HEM) guarantees
our results depend on \citep{aarne2024, ogara2025}. \Cref{sec:hems} discusses
the HEM requirements in more detail as a limitation and as a direction for
development.

The hardware operator is theoretically more challenging to monitor than the
cloud tenant, but both evaders are subject to a fundamental physical constraint:
a backward pass requires $\sim$2$\times$ the FLOPs of a forward pass, and
therefore flips more transistors and consumes more energy. Evasion strategies
designed against one category of evader transfer reasonably well to the other,
because one major constraint is the physical energy requirement, not specific
software.

\subsection{Classification Task}
\label{sec:task}

Our dataset contains telemetry records from our 93 distinct ML training
workloads (34 ordinary ML tasks, 59 adversarial workloads disguised as inference
or non-ML), 40 distinct ML inference workloads, and 19 non-ML workloads. Our
classifier can distinguish these with high accuracy and precision, as described
in \cref{sec:methods,sec:results} below. But the fundamental governance-relevant
question is binary: \textbf{Is a monitored GPU being used for training?} We
tested several options before selecting a random forest classifier, which
exhibited the best performance on this task during cross-validation. Details are
provided in \cref{app:classifier-design}.

\section{Related Work}
\label{sec:related}

The field of compute governance contains many proposals for regulations that
implicitly rely on the ability to reliably identify when developers are using
compute for AI training, and many proposals for hardware-enabled mechanisms that
assist our classifier in securely providing that information.
\citet{shavit2023} proposes compute monitoring for verifying rules on
large-scale training. \citet{sastry2024} provide a comprehensive framework for
compute as a governance lever. \citet{heim2024thresholds} analyse FLOP
thresholds as a regulatory instrument; for example, the $10^{26}$-FLOP line in
U.S.\ EO 14110 is a direct operationalisation of this. \citet{scher2025} propose
an international agreement wherein training above certain compute thresholds is
monitored and restricted or prohibited, while inference is allowed; this would
require the ability to distinguish between inference and training on monitored
hardware. \citet{baker2025} identify six verification layers for international
AI agreements; our work provides empirical evidence for one layer (built-in
monitoring). \citet{ogara2025} and the RAND workshop report \citep{kulp2025}
evaluate HEM options including workload attestation. \citet{petrie2024} proposes
firmware-based offline licensing for export controls. \citet{seferis2025} argue
that compute-structuring evasion (splitting training across providers) is
detectable conditional on per-host workload classification; we supply and
stress-test exactly that per-host classifier. \citet{diffarch2026} propose
hardware mechanisms to limit performance of targeted applications, a type of
enforcement complementary to our detection approach. One possibility is to
robustly classify all workloads on hardware within the governance jurisdiction,
and then apply the differential-architecture performance restrictions to
automatically enforce the established rules.

There is also existing research relevant to workload classification, which we
extend. \citet{monfared2026} study timing and memory observables for detecting
training when host and device are untrusted, constructing their own derived
signals from primitives that survive a strong tampering adversary. They
establish signals that remain informative when NVML cannot be relied upon. Our
case is complementary, establishing highly successful classification from
signals that will be reliable if they can be protected from tampering by HEMs.
Their work establishes a lower floor on the signal set; ours establishes an
upper bound on what an adversary can hide inside that signal set.
\citet{chen2025} develop a hardware-centric, workload-agnostic approach to
hardware monitoring, analyzing over 30 ML models on various hardware platforms
to perform anomaly detection using an unsupervised learning pipeline. Our scope
includes supervised examples, using more than 1400 labeled records of workloads
totaling over 100 hours of hardware signal telemetry to teach the classifier to
recognize ML training from those signals. \citet{gangwal2019} investigate a
conceptually very similar problem, using hardware performance counters for
workload classification using RF and SVM models, but with the objective of
discovering covert cryptomining, not ML training. The study notes an adversarial
limitation: they predicted that evasion may be possible by restricting mining
operations to obfuscate the workload signature, but at a profitability cost,
analogous to the throughput penalty that evaders would suffer while attempting
to evade classifiers such as ours. We extend the workload classification
objective to the adversarial setting, where operators attempt to deceive the
classifier.

Other past work covers several adjacent but distinct tasks. WAVE
\citep{xu2026} identify inference workloads with 87--93\% accuracy using 24
Nsight Compute per-kernel counters (1200--5300\% overhead). We accomplish a
similar-granularity classification using only always-available NVML physical
counters at zero runtime overhead; the tasks are not identical (WAVE separates
inference families; we separate training from inference and others), so results
are not directly comparable. \citet{esposito2025} combine NVML telemetry and
performance counters to estimate GPU stress and reliability across workloads.
Their results confirm that telemetry reveals meaningful workload differences,
but the target is reliability estimation rather than governance-oriented
training detection. ShadowScope \citep{almusaddar2025} uses composable
performance-counter signals to detect anomalous behaviors and prevent attacks
which could compromise the integrity of GPU computation. We collect and analyze
similar data, but using NVML readings rather than CUDA profiling events.

Using physical side channels rather than software telemetry, \citet{maia2022}
show magnetic flux from GPU power cables reveals neural network topology.
\citet{arefin2025} demonstrate USB/HDMI port power measurement can infer GPU
activities. \citet{latif2025} empirically measure power signatures of training
on 8-GPU H100 nodes. These establish that physical signals carry
workload-identifying information; we extend this to software-accessible
telemetry for governance. POLCA \citep{patel2023} characterize LLM power
consumption profiles across a Microsoft Azure fleet for energy planning and find
systematic differences in power patterns between training and inference,
providing independent evidence for the signal detected by our classifier, at the
scale of large cloud datacenters.

\section{Methodology}
\label{sec:methods}

\subsection{Data Collection}
\label{sec:data}

\subsubsection{Hardware}

Hardware used for this project includes 9 models of NVIDIA GPUs across 4
architecture generations. Consumer hardware is generally not covered by proposed
AI governance and verification \citep{heim2024cloud, scher2025}, but was included
to test generalization across different hardware.

\begin{table}[t]
\caption{Instances used in this research by type and quantity of GPUs. Some
experiments involved fewer than the maximum number of GPUs available on the
instance. NVML distinguishes ``H200'' and ``H200 NVL'' only by marketing
name---the two SKUs share identical 141\,GB HBM3e and identical telemetry
signatures---so we collapse them into a single H200 class; H100 80GB HBM3 and
H100 NVL are kept separate.}
\label{tab:hardware}
\vskip 0.10in
\begin{center}
\begin{small}
\setlength{\tabcolsep}{4pt}
\begin{tabular}{@{}llccc@{}}
\toprule
GPU Model        & Architecture  & VRAM   & Context        & GPUs \\
\midrule
RTX 4050 Laptop  & Ada Lovelace  & 6\,GB   & Laptop  & 1 \\
RTX 5080         & Blackwell     & 16\,GB  & Desktop & 1 \\
RTX 5090         & Blackwell     & 32\,GB  & Cloud         & 1, 2 \\
A100 SXM4 40GB   & Ampere        & 40\,GB  & Cloud         & 1, 8 \\
A100 SXM4 80GB   & Ampere        & 80\,GB  & Cloud         & 1, 8 \\
H100 80GB        & Hopper        & 80\,GB  & Cloud         & 1, 8 \\
H100 NVL         & Hopper        & 94\,GB  & Cloud         & 1, 2 \\
H200 / H200 NVL  & Hopper        & 141\,GB & Cloud         & 1, 2, 8 \\
B200             & Blackwell     & 192\,GB & Cloud         & 1, 2, 8 \\
\bottomrule
\end{tabular}
\end{small}
\end{center}
\vskip -0.1in
\end{table}

\subsubsection{Workloads}

We conducted 1404 runs of 162 distinct workloads, totaling $\sim$117 GPU-hours
($\sim$65 wall-clock hours) of experiments, for a total of 445{,}310 telemetry
readings $\times$ 9 signals per reading. Not every workload is run on every
hardware configuration, for instance due to memory limitations.

\begin{table}[t]
\caption{Number of workloads and runs in each of the three overall categories.}
\label{tab:workloads}
\vskip 0.10in
\begin{center}
\begin{small}
\begin{tabular}{@{}lcc@{}}
\toprule
Category     & Workloads & Runs \\
\midrule
ML training  & 93 & 867 \\
ML inference & 40 & 352 \\
Other        & 16 & 185 \\
\bottomrule
\end{tabular}
\end{small}
\end{center}
\vskip -0.1in
\end{table}

We conducted 867 runs of ML training, including 180 of supervised training
(e.g.\ ResNet \citep{he2016} on CIFAR-10, GPT-2 on WikiText, BERT
\citep{devlin2019} fine-tuning), 72 of distributed data-parallel training, 152
of fully-sharded data-parallel training on models with 7B to 70B parameters, and
42 of LoRA fine-tuning. This also included 421 runs of adversarial evasion,
testing strategies such as utilization modulation, memory minimization or
preallocation, clock throttling, interleaving training with inference or non-ML
workloads, mimicking FFT computations or cryptocurrency mining, and padding with
idle time.

For inference, we did 352 runs, including 49 of image classification, 264 of LLM
inference (TinyLlama, Phi-3, Qwen-2.5 3B through 72B, Mistral and Mixtral,
DeepSeek-V3 and R1), and 39 of other modalities (e.g.\ Whisper speech-to-text,
SDXL text-to-image diffusion).

We also did 185 runs of non-ML workloads, including 65 of scientific HPC (e.g.\
fast Fourier transform, n-body simulation), 36 of cryptocurrency mining, 36 of
rendering or video processing (Blender, rendering, FFMPEG), 34 idle-GPU
baselines, and 14 miscellaneous non-ML runs.

A full catalog of adversarial workloads is provided in
\cref{app:strategy-catalog} and source code for every workload is available in
the project repository.

\subsubsection{Telemetry}

During all workloads, we sample 9 raw NVML signals at 1\,Hz: GPU utilization
(\%), memory utilization (\%), memory used (MB), power draw (W), temperature
($^\circ$C), SM clock (MHz), memory clock (MHz), PCIe TX bandwidth (MB/s), and
PCIe RX bandwidth (MB/s).

Additionally, for 49 workloads tested on the local RTX 5080 desktop, we validate
the telemetry's reported power draw using an external Tapo P115 smart plug
reading the system's power draw with timestamps matching within 0.5\,s. We
extract features from the raw signals as described in \cref{app:features}.

\subsection{Classifier}
\label{sec:classifier}

We selected the best-performing of several candidate models. RandomForest
\citep{breiman2001} with 400 trees and no maximum depth was the most accurate,
and XGBoost \citep{chen2016xgboost} was a close runner-up. We use RF for all
reported headline results but retain XGBoost for gain-based feature-importance
analysis (\cref{fig:lood}). SVM-RBF and logistic regression were also
considered, but exhibited worse performance during cross-validation. Classifier
comparison is provided in \cref{app:classifier-selection}.

The feature set is designed along two signal layers: (1) a structural, roofline
\citep{williams2009} layer that identifies workloads occupying a training-like
operating region in arithmetic-intensity space (mean power, mean utilization),
and (2) a temporal telemetry layer that resolves ambiguity within that region
using time-domain features (autocorrelation, CV, periodicity). Both stages of
the two-stage pipeline use the same underlying feature vocabulary: stage 1
retains the full set; stage 2 drops the level features to reduce overfitting and
prevent exploitation by adversarial strategies that mimic specific usage levels.

\subsection{Adversarial Evaluation Protocol}
\label{sec:protocol}

Our refinement procedure iterates on the classifier and the evader through five
rounds of co-evolution. In each round, we design new evasion strategies for the
attacker, targeting the classifier's existing weaknesses, and then the defender
trains a hardened classifier incorporating the new adversarial examples.
Qualitatively, the classifier improves through stages which can be described as
\emph{baseline}---trained only on non-adversarial data consisting of real-life
workloads---to \emph{hardened}---defeating initial adversarial strategies
introduced in earlier rounds---to \emph{evasion-resistant}---hardened against
more sophisticated obfuscation, plus sensing physics-based features that capture
fundamental computational invariants that the adversary cannot suppress without
aborting training.

\begin{table*}[t]
\caption{Summary of adversarial iteration over several rounds of increasing
evader and classifier sophistication. Workload labels are in parentheses; a
full catalog is provided in \cref{app:strategy-catalog}.}
\label{tab:rounds}
\vskip 0.10in
\begin{center}
\begin{small}
\begin{tabular}{@{}p{0.9cm}p{6.9cm}p{7.1cm}@{}}
\toprule
Round & Attacker introduces & Defender response \\
\midrule
1 & Utilization manipulation: modulation (A), low-utilization training (B),
    temporal disruptions (D) & Retrain $\rightarrow$ \textbf{hardened}
    classifier \\
2 & Memory reduction and non-ML mimicry: minimal-memory training (E),
    interleaving (F), clock throttling (G), FFT/mining mimicry (H), stochastic
    (I), PID control (J) & Retrain $\rightarrow$ \textbf{hardened} (improved) \\
3 & Training-as-inference: online learning (K), training diluted with
    $N \in \{2,5,10,20\}\times$ inference (L) & Add power--memory correlation
    $+$ cumulative energy features $\rightarrow$ \textbf{evasion-resistant} \\
4 & Composite evasion: min-memory composite strategy (M), gradient accumulation
    structuring (N), composite idle-padding sweep (0--95\% idle fraction);
    multi-GPU distributed data-parallel variants (K\_ddp, K\_ddp\_\allowbreak accum, L\_ddp,
    L\_ddp\_\allowbreak stagger) & Retrain on multi-GPU data $\rightarrow$
    \textbf{evasion-resistant} (improved) \\
5 & White-box (attacker given classifier internals): diluted full-model
    fine-tuning (WB-D), diluted LoRA fine-tuning (WB-L), WB $+$ grad-accum
    combined & Add pre-allocation memory-plateau features
    (\texttt{first\_\allowbreak 30s\_\allowbreak mem\_\allowbreak delta}, \texttt{time\_\allowbreak to\_\allowbreak mem\_\allowbreak plateau}); collect
    more non-adversarial examples of matched-size dense LLM inference on
    datacenter hardware \\
\bottomrule
\end{tabular}
\end{small}
\end{center}
\vskip -0.1in
\end{table*}

This iteration process yielded 20 families of evasion strategies covering six
technique classes: utilization manipulation, memory reduction, workload
mimicry, training-as-inference (e.g.\ dilution or mixing the two), multi-GPU
evasion (e.g.\ structuring workloads across multiple GPUs to look less like
training on some of them individually), and white-box adversary (examining
previous classifiers to discover niche weaknesses).

For each strategy family, we train the classifier on real (non-adversarial)
training, inference, and non-ML workload data, plus every \emph{other}
adversarial strategy, and test on the held-out family. This tests the
classifier's ability to generalize to unseen, novel evasion techniques, which
informs us about the real-world governance case where an adversary invents a
strategy we have not yet catalogued.

\section{Results}
\label{sec:results}

\subsection{Baseline Classification}
\label{sec:baseline}

\begin{figure}[t]
\vskip 0.1in
\begin{center}
\centerline{\includegraphics[width=\columnwidth]{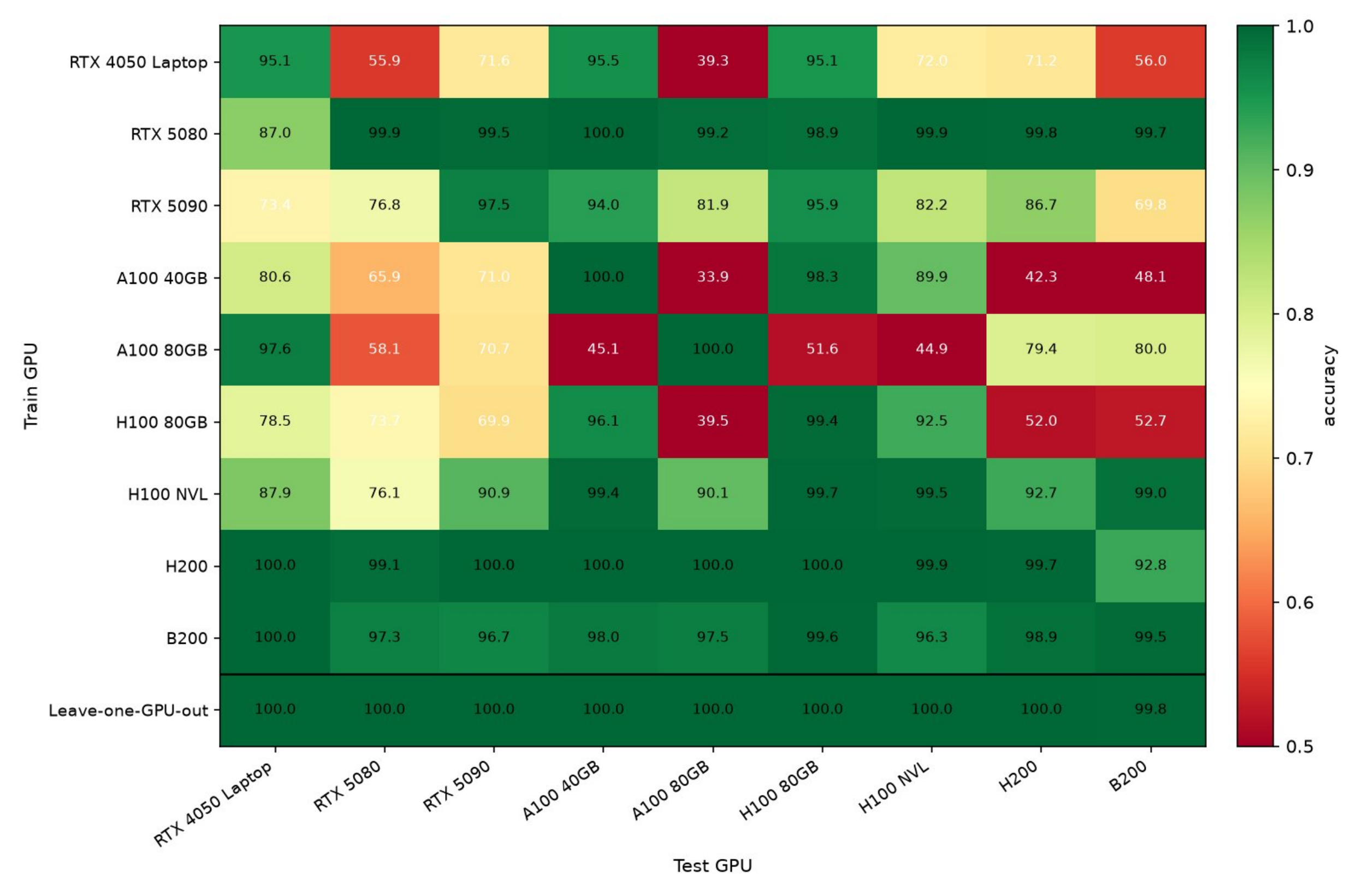}}
\caption{Cross-GPU transfer heatmap (binary RF, 30\,s windows). Cell values are
accuracy when training on the row GPU and testing on the column GPU.
Datacenter-to-datacenter transfer is consistently above 90\%. Cross-consumer and
consumer$\leftrightarrow$datacenter transfer is weaker; RTX 4050/5080/5090 rows
often drop below 50\% on datacenter targets. The bottom row is LOGO: for each
test GPU, the classifier is trained only on data from the other 8 GPUs. By
contrast, \emph{in-distribution} evaluation (with runs from the target GPU
included in the training dataset) reaches 100\% on unseen non-adversarial runs,
for every GPU. LOGO evaluates the harder case of generalization to new hardware
types.}
\label{fig:transfer}
\end{center}
\vskip -0.2in
\end{figure}

First we present the non-adversarial, pre-hardening baseline. We replicate the
previous finding that the classification problem is solvable and real training
and inference workloads can be distinguished from non-ML workloads
\citep{tang2022, xu2026} such as cryptocurrency mining and graphics rendering.

Pooling all 9 GPU models, the classifier achieves 98.21\% $\pm$ 0.84\% accuracy
with macro-F1 $= 0.982$.\footnote{RandomForest, 30\,s windows, stride 15\,s,
5-fold grouped CV, 26{,}496 windows across 1{,}396 runs that produced at least 5
NVML samples.} Predicting classes of held-out runs, it achieves 97.39\%
precision and 98.90\% recall, with 96.94\% specificity and 3.06\% false-positive
rate. Latency is quite quick, with 98.15\% accuracy at 5\,s windows and 98.21\%
at 30\,s. Analysis of detection latency is provided in \cref{app:window}.
Because the classifier emits a calibrated $P(\text{training})$, the decision
threshold can be adjusted depending on the context, without retraining. For
example, a regulator monitoring a facility for covert training would prefer high
sensitivity, catching every training run even at the cost of more false
positives, whereas a compute provider would likely prefer to use the classifier
with a higher training threshold to avoid flagging legitimate customer
workloads.

Accuracy was evaluated with the data split by GPU model, comparing single-GPU
training against training on data pooled across all 9 GPUs.

\begin{table}[t]
\caption{Classification accuracy by GPU. The Single-GPU CV column trains and
tests on different workloads from the same GPU using 5-fold grouped
cross-validation. Runs are randomly allocated to a single fold; all windows
from the same run are within the same fold, so data from a run never appears in
both train and test data. The LOGO (leave-one-GPU-out) column trains on data
from the \emph{other} 8 GPUs and tests on the held-out GPU.}
\label{tab:bygpu}
\vskip 0.10in
\begin{center}
\begin{small}
\begin{tabular}{@{}lcc@{}}
\toprule
GPU              & Single-GPU CV & LOGO \\
\midrule
A100-SXM4 40GB   & 100.00\% & 100.00\% \\
A100-SXM4 80GB   & 100.00\% & 100.00\% \\
B200             & 99.55\%  & 99.80\% \\
H100 NVL         & 99.52\%  & 100.00\% \\
H200             & 99.73\%  & 100.00\% \\
H100 80GB HBM3   & 99.43\%  & 100.00\% \\
RTX 5090         & 97.57\%  & 100.00\% \\
RTX 5080         & 99.93\%  & 100.00\% \\
RTX 4050 Laptop  & 95.14\%  & 100.00\% \\
\bottomrule
\end{tabular}
\end{small}
\end{center}
\vskip -0.1in
\end{table}

In-distribution pooled classification is essentially solved for non-adversarial
workloads: when data from all 9 GPUs is used to develop the classifier, it is
100\% accurate on new, non-disguised workloads performed on any of those GPUs. A
harder case is when the classifier is developed using workload logs on a set of
GPUs and then it is deployed to classify telemetry from a novel GPU. This case
is evaluated in the LOGO column of \cref{tab:bygpu}, which shows that the
classifier generalizes well, reaching 100\% on 8 out of 9 GPUs and at worst
99.4\% on the B200, despite not being trained on any data from the target
hardware. Cross-GPU transfer when training on data from a \emph{single} GPU
(illustrated in the upper rows of \cref{fig:transfer}) is harder, especially
when training on a consumer GPU and testing on a datacenter GPU, but is not
relevant for governance. Our results show that a regulator with access to
telemetry from a variety of GPUs can train one classifier and deploy it
site-agnostically, even to GPU models they have not sampled before.

\subsection{Adversary--Defender Iteration}
\label{sec:armsrace}

\Cref{tab:armsrace} shows the key results. The first row shows baseline
classification performance on only clean, non-adversarial workloads. The
baseline classifier, trained on all but one workload at a time, achieves an
average of 97.6\% accuracy on the other, held-out workload.

\begin{table*}[t]
\caption{Arms-race summary showing each classifier state's detection rate against
each round's evasion strategies. Detection rate for each strategy is computed
over runs that employ that strategy; workloads not affected by the strategy are
not included. Baseline classifiers are trained only on non-adversarial data. All
classifiers use class-balanced training, with class weights adjusted inversely
to their number of samples.}
\label{tab:armsrace}
\vskip 0.10in
\begin{center}
\begin{small}
\begin{tabular}{@{}p{1.4cm}p{4.6cm}cccc@{}}
\toprule
Round & Attacker capability & Baseline det.\ & Prev.-round det.\ & Hardened
(in-dist.) & Hardened (unseen) \\
\midrule
Baseline & (No evasion; real workloads only) & N/A & --- & --- &
   \textbf{97.6\%}\,$^{a}$ \\
1 & Utilization manipulation (A, B, D) & 6--96\%\,$^{b}$ & --- & 100\% &
   \textbf{59--100\%} \\
2 & Memory reduction / mimicry (E, F, G, H, I, J) & 0--91\% & 20--95\% & 100\% &
   \textbf{80--100\%} \\
3 & Training-as-inference (K online, L diluted) & 52--60\% & 48--57\% & 100\% &
   \textbf{87--100\%} \\
4 (single-GPU) & Composite $+$ grad accum (M, N, composite 0--95\% idle) &
   19--100\% & 83--100\% & 100\% & \textbf{98--100\%} \\
4 (multi-GPU) & Multi-GPU distributed evasion (K/L DDP $+$ stagger) & 72--100\% &
   71--100\% & 100\% & \textbf{100\%} \\
5 & White-box (WB-full, WB-LoRA) with R4 internals exposed & N/A\,$^{c}$ &
   6--100\% & 100\% & \textbf{43--87\%} \\
\bottomrule
\end{tabular}
\end{small}
\end{center}
\footnotesize
$^{a}$ Prior to Round 1, this refers to the baseline classifier's accuracy at
evaluating held-out non-adversarial workloads when trained on all other
non-adversarial workloads. $^{b}$ Prior to Round 2, there was no hardened
classifier; the previous round's classifier was the baseline classifier trained
only on non-adversarial workloads. $^{c}$ The white-box evader in Round 5 has
access to the evasion-resistant classifier produced in Round 4. If evaluated
against the \emph{baseline} classifier with access to its internals, detection
would be very low, likely 0--20\%.
\normalsize
\vskip -0.1in
\end{table*}

In Round 1, the evader introduces workload obfuscation strategies revolving
around utilization manipulation. The baseline classifier achieves 6--96\% binary
accuracy at classifying workloads using these strategies as training vs.\
not-training. Then the defender iterates on these strategies, achieving 100\%
in-distribution accuracy at classifying these workloads after it has seen them,
or 59--100\% accuracy at classifying unseen strategies from Round 1 after
learning from the other strategies.

In Round 2, the evader introduces memory reduction and mimicry strategies, and
in Round 3, the evader introduces so-called ``training-as-inference''
strategies, camouflaging online learning as inference or diluting gradient
updates between many batches of inference. Each time, the evader's new
strategies achieve partial success against the baseline classifier and the
previous round's classifier, and each time, the defender trains the new
classifier to always detect the new evasive workload if it has seen that
strategy, or detect them the majority of the time if it has seen strategies from
that round but not the specific strategy it is tested on.

In Round 4, we introduced new evasion strategies for disguising single-GPU
training, as well as strategies for disguising multi-GPU training which attempt
to prevent the detection of training on each GPU's telemetry by utilizing other
GPUs to shift the workload around. Round 4's strategies achieved some initial
evasion against the classifier from Round 3, but were very thoroughly rendered
ineffective once the classifier was hardened against them.

Finally, in Round 5, we gave the evader white-box access to the classifier
developed in previous rounds. The evader was able to examine the internal
mechanics of the classifier to find precise exploits. For example, while
red-teaming the Round 4 classifier, we noted that the classifier partly relies
on features derived from windows 5\,s or longer. Therefore, one prospective
evader strategy is to insert 4\,s of training across the boundary of two 5\,s
windows, spending 40\% of 10\,s on training but taking up less than half of the
time of each window, misleading the classifier towards identifying this as the
other workload that was running. Another tactic was to use a copy of the Round 4
classifier against the defender in real-time, running training as long as the
defender believed $P(\text{training}) < 0.30$ and temporarily replacing training
with another workload whenever the defender became suspicious. These tactics
achieved up to 94\% evasion against the Round 4 classifier. However, once the
defender was able to train against them, they were rendered ineffective, with
(as usual) 100\% detection of seen evasion strategies, and notably 43--87\%
detection of unseen white-box strategies after hardening against other Round 5
strategies.

Round 5 represents a disaster scenario for governance, equivalent to a developer
infiltrating the governing agency, stealing the classifier, and orchestrating
their workload in a way that is designed to be maximally undetectable by the
classifier's exact algorithm.\footnote{In an absolute worst-case scenario, the
developer could infiltrate the regulator, replace the classifier with a
compromised version containing a backdoor, and also compromise the monitored
hardware.} However, once a classifier is hardened against a few such exploits,
other white-box attacks are much less effective in the future.

\section{Discussion}
\label{sec:discussion}

\subsection{An Involuntary Witness}
\label{sec:witness}

During ML training, a backward pass (computing gradients at each parameter)
involves twice as much computation as the forward pass \citep{kaplan2020}. This
imposes a fundamental constraint on training: for the same number of tokens
processed, it is three times as energy-intensive as inference, so you can only
either shuffle energy consumption around or reduce the amount of training
conducted, but ultimately you cannot make them use the same amount of energy
without eliminating the step of updating the weights, which would leave the
workload unable to train the model.

Our empirical evidence is consistent with this and suggests that power draw is a
valuable feature for distinguishing training from
inference.\footnote{This theoretical result is corroborated by our experiments
as well as evidence from large-scale training and inference deployments. For
example, POLCA \citep{patel2023} find that inference uses less power than
training in Microsoft Azure cloud datacenters.} Speculatively, as we iterate
further on the classifier--evader process in the future, evaders may learn to
defeat other features, but energy is a fundamental constraint, so energy-based
features may rise in importance as the classifier hardens further.

\subsection{Single- vs.\ Multi-GPU Detection}
\label{sec:multigpu}

Our datacenter experiments (on 8-GPU A100, H200, B200 nodes) test both threat
models above. We designed 4 multi-GPU evasion strategies: DDP online learning,
DDP diluted training, staggered dilution (spreading NVLink traffic across time),
and gradient accumulation across ranks. All are detected at 99.6--100\% in
leave-one-strategy-out evaluation (\cref{tab:strategies}; \cref{fig:lood}).
Multi-GPU training is harder to hide than the much less important single-GPU
case because traffic and interconnect usage patterns that differ between
ordinary multi-GPU large-model training and inference provide additional
opportunities for the classifier to detect differences in signals, and more that
the evader will have to convincingly disguise. For example, our classifier
learned to recognize AllGather and ReduceScatter NVLink traffic associated with
forward/backward steps in FSDP training at 7B--70B scale and applied this to
detect obfuscated multi-GPU training even when the adversary tried to manipulate
the communication patterns. Multi-GPU training may create distinctive telemetry
signatures on each GPU even when workloads are structured so that an individual
GPU shows only moderate utilization.

\subsection{Privacy Preservation}
\label{sec:privacy}

The NVML metrics queried by our classifier are content-agnostic: it observes the
physical and architectural effects of computation (power draw, temperature,
utilization, memory) without accessing model weights, training data,
hyperparameters, or application logic. This addresses a central deployment
concern in shared cloud settings, where governance mechanisms that require
inspecting proprietary information face legal, operational, and ethical
objections \citep{shavit2023, ogara2025}. Our classifier demonstrates that the
governance-relevant question (``is this GPU training a model?'') can be answered
without any access to the semantic content being processed.

\subsection{Layered Defense}
\label{sec:layered}

If telemetry is collected and monitored at multiple layers (device, cluster,
facility), an adversary must evade all layers simultaneously \citep{ogara2025}.
The evasion cost is cumulative across layers, so our results using NVML telemetry
are a lower bound on what a layered monitor can achieve. DCGM counters
(tensor-core activity, NVLink throughput) and wall- or rack-level power
measurements provide additional signal. If combined with tamper-resistant
hardware such as flexHEGs \citep{petrie2025flexheg}, software-level evasion alone
could not defeat such a layered monitor.

\subsection{Implications for HEMs}
\label{sec:hems}

Our results naturally suggest a minimal HEM design for preventing covert training
on monitored hardware. The software-level classifier demonstrates that
always-available NVML counters at 1\,Hz carry enough signal to detect
training-versus-not at 98.2\% accuracy with leave-one-out adversarial
robustness---but this result only holds if the counters themselves are
trustworthy. A deployable monitoring system therefore requires three HEM
components in addition to the classifier:

\begin{enumerate}
\item \textbf{Tamper-resistant NVML read path.} The monitoring agent must read
counter values from a source the adversary cannot tamper with---e.g., an
attested enclave on the GPU firmware boundary \citep{aarne2024} or a verified
read path through secure-boot-chained firmware \citep{petrie2024, ogara2025}.
Without this, an adversary with software privilege simply returns forged values.
\item \textbf{Authenticated telemetry channel.} Telemetry in transit from the
GPU agent to the verifier must be signed with a hardware-rooted key so an
in-path adversary cannot substitute a permitted workload's telemetry for the
actual workload. This extends to rate-limiting and nonce-binding to prevent
replay of previously-observed benign signatures.
\item \textbf{Secure boot of the monitoring software.} The process itself, its
classifier weights, and its feature-extraction code must be attested so an
adversary cannot replace them with a permissive version (e.g., a classifier that
labels everything as inference).
\end{enumerate}

\section{Conclusion}
\label{sec:conclusion}

Zero-overhead GPU telemetry monitoring is a viable mechanism for compute
governance. Defenders are structurally favored, especially in the most relevant
case: our classifier is extremely effective at detecting all evasion strategies
tested on multi-GPU datacenter nodes, correctly classifying 98--100\% of
unforeseen workloads in that setting. Single-GPU evasion has slight success, with
a 6.2\% evasion rate in held-out evaluation, on average across the 14 single-GPU
non-white-box strategies. Only one type of evasion, white-box LoRA, achieves
under 50\% detection rate against the evasion-resistant classifier, even when
evaders have access to the defender's previous classifier. The classifier learns
to recognize fundamental features that cannot be eliminated without sacrificing
the ability to update model weights. The results are robust across four
generations of NVIDIA hardware, with strong generalization to other GPU models
in hold-out validation.

The compute-monitoring proposals of \citet{shavit2023}, \citet{baker2025}, and
\citet{ogara2025} are empirically realisable against a realistic software-level
evasion adversary, conditional on the HEM assumptions of tamper-resistant
telemetry, an authenticated telemetry channel, and a secure classifier. With
those hardware guarantees in place, the adversary is left with a physical cost
trade-off (dilution and idle-padding) whose economics are unviable at frontier
scale. By demonstrating that it is possible to construct a classifier that
accurately identifies training activity, we advance the progress of governance
from the ML problem to the hardware security problem: ``Can we build the matching
HEM?'' We invite researchers and ML experts to red-team our classifier and
publish evasion strategies that might defeat it, so that we can iterate on any
such findings to ultimately strengthen its robustness.

\section*{Impact Statement}

Key to many AI governance proposals is the ability to narrowly detect and
restrict dangerous frontier AI training. Governance solutions are more likely to
be adopted by governments and regulators if they are robust against evasion and
preserve the privacy of law-abiding monitored parties. Our classifier
establishes the state of the art in robust compute workload monitoring
strategies, an essential ingredient in making minimally invasive options
available to regulators seeking to govern future AI development.

\bibliography{references}
\bibliographystyle{icml2022}

\newpage
\appendix
\onecolumn

\section{Classifier Design}
\label{app:classifier-design}

\subsection{Deployed Classifier}
\label{app:deployed}

The operational form we deploy is a two-stage pipeline that factors the decision
along a natural tree:

\begin{enumerate}
\item \textbf{Is the workload ML?} A binary RandomForest head using all features
extracted from telemetry readings classifies the workload into
$\text{ml\_training} \cup \text{ml\_inference}$ vs.\ other. The ML-vs-non-ML
boundary leans heavily on level signals (peak memory footprint, sustained
tensor-core utilization, power draw), so the full feature set is retained.
\item \textbf{If the workload is ML, is it training?} A binary RandomForest head
trained only on records where the true category is training or inference
classifies incoming readings into either category. This head receives a subset
of features with level features dropped (absolute memory/power/clock means,
percentiles, min, max, range). The train-vs-infer decision rests on shape,
periodicity, FFT, and autocorrelation terms only. This prevents the classifier
from overfitting to specific models previously tested and specific power, memory,
or compute throughput levels associated with particular GPUs. This also makes
the classifier more robust to certain adversarial strategies: for example, the
evader cannot modulate their usage levels to match that of another chip that is
specifically designed for inference, because the classifier does not see the
absolute levels, only the utilization ratios and temporal patterns.
\end{enumerate}

The two-stage form matches the governance decision surface, makes both stages
independently auditable and interpretable, and avoids the vulnerability where a
single multi-class head is uncertain between training/inference, splits
probability between them, and labels a non-ML workload as most probable.

\subsection{Classifier Selection}
\label{app:classifier-selection}

We evaluated several types of models and selected the candidate that performed
best in holdout cross-validation testing. The strongest were RandomForest
\citep{breiman2001} (\texttt{n\_estimators=400}, \texttt{max\_depth=None},
\texttt{min\_samples\_leaf=2}, \texttt{max\_features="sqrt"},
\texttt{class\_weight="balanced"}) and XGBoost (\texttt{n\_estimators=500},
\texttt{max\_depth=6}, \texttt{lr=0.05}, sample weights via
effective-number-of-samples class balancing), which we evaluated head-to-head on
15\,s, 30\,s, and 60\,s windows with 5-fold cross-validation. RandomForest wins
by a small margin at every window size; at 30\,s it reaches 98.21\% $\pm$ 0.84\%
accuracy / 0.982 macro-F1. We report RF 30\,s as the headline operating point,
but note that RF and XGB perform similarly well, so the main findings of this
study do not depend on the particular type of classifier selected.

We also tested SVM-RBF (RBF kernel, standardised features, balanced class
weights) and Logistic Regression (saga solver, L2 regularisation, standardised
features).

\begin{table}[h]
\caption{Performance of each type of classifier at identifying held-out
workloads during cross-validation.}
\label{tab:classifiers}
\vskip 0.10in
\begin{center}
\begin{small}
\begin{tabular}{@{}lcc@{}}
\toprule
Classifier        & Accuracy              & Macro-F1 \\
\midrule
RandomForest      & \textbf{98.21\% $\pm$ 0.84\%} & \textbf{0.982} \\
XGBoost           & 98.14\% $\pm$ 0.78\%  & 0.981 \\
SVM-RBF           & 96.19\% $\pm$ 0.89\%  & 0.962 \\
Logistic Reg.     & 89.81\% $\pm$ 0.83\%  & 0.897 \\
\bottomrule
\end{tabular}
\end{small}
\end{center}
\vskip -0.1in
\end{table}

The tree ensembles dominate by $\sim$2\,pp over SVM-RBF and $\sim$8\,pp over
logistic regression, which is consistent with the feature set carrying
non-linear interactions that a kernel-SVM or linear classifier cannot exploit as
cheaply as a tree ensemble. SVM-RBF at 96.2\% is a useful cross-check that the
tree-ensemble headline is not an artifact of the specific algorithm family. The
drop to 89.8\% for logistic regression quantifies the non-linearity of the
problem: a purely linear decision boundary in the 166-feature space is
$\sim$8\,pp worse, confirming that the classifier is exploiting feature
interactions (power CV $\times$ memory slope, autocorrelation $\times$ FFT
periodicity, etc.) not just per-feature thresholds.

\subsection{Feature Extraction}
\label{app:features}

We take 5\,s, 15\,s, 30\,s, and 60\,s sliding-window snapshots of telemetry
readings and extract 166 features per window. The composition is:

\begin{itemize}
\item Base statistics (9 signals $\times$ 13 stats $=$ 117 features): mean, std,
min, max, percentiles (25/50/75/95), IQR, range, CV, skew, kurtosis.
\item Autocorrelation (3 signals $\times$ 5 lags $=$ 15 features): lags
1/2/5/10/20 for utilization, power, memory.
\item Cross-signal features: power/utilization ratio, PCIe total, utilization
per SM\%.
\item Training-specific temporal features: memory slope, epoch periodicity (via
FFT of power).
\item Memory-plateau features: absolute and percentage change in memory usage in
first 30\,s, and time to memory plateau. These were motivated by an adversarial
strategy that learned to spoof an earlier classifier by pre-allocating memory
before it was needed for training, enabling it to hide memory allocation during
training.
\item Inference-specific features: duty cycle, idle fraction, utilization
transitions.
\end{itemize}

\subsection{Adversarial Strategy Catalog}
\label{app:strategy-catalog}

Families are ordered by round. Baseline and Hardened columns report the binary
classification accuracy (or accuracy range, for strategies with variants). The
LOO column reports the accuracy (or mean accuracy, for strategies with variants)
when training the classifier on evasion strategies from the same and earlier
rounds, but holding that row's strategy out of the training data.

\begin{table}[h]
\caption{The classifier's detection rates against each evasion family. Bolded
baseline cells mark families with at least one variant detected at less than
90\%. Bolded LOO cells mark family means detected at less than 90\%. The hardest
group of strategies for our classifier is \texttt{whitebox\_lora}: LoRA's small
footprint, editing only a few parameters, combined with heavy dilution strips the
training signature to near the noise floor. All R1--R4 families are detected at
100\% in the in-distribution hardened state.}
\label{tab:strategies}
\vskip 0.10in
\begin{center}
\begin{small}
\begin{tabular}{@{}lllcccc@{}}
\toprule
Family & Round & Technique & Baseline & Hardened & LOO mean & Variants \\
\midrule
\texttt{A\_util\_modulation}   & 1 & Utilization manipulation & \textbf{6\%}    & 100\% & 93.9\%          & 1 \\
\texttt{B\_low\_util}          & 1 & Utilization manipulation & \textbf{11\%}   & 100\% & \textbf{59.1\%} & 1 \\
\texttt{D\_temporal\_disruption} & 1 & Utilization manipulation & 96\%        & 100\% & 100\%           & 1 \\
\texttt{E\_memory\_minimal}    & 2 & Memory reduction         & \textbf{18\%}   & 100\% & 98.2\%          & 1 \\
\texttt{F\_interleave}         & 2 & Workload mimicry         & \textbf{34--36\%} & 100\% & 100\%         & 2 \\
\texttt{G\_clock\_throttled}   & 2 & Utilization manipulation & \textbf{36\%}   & 100\% & 98.2\%          & 1 \\
\texttt{H\_mimicry}            & 2 & Workload mimicry         & \textbf{18--27\%} & 100\% & 99.1\%        & 2 \\
\texttt{I\_stochastic}         & 2 & Utilization manipulation & 91\%            & 100\% & 100\%           & 1 \\
\texttt{J\_pid}                & 2 & Utilization manipulation & \textbf{0\%}    & 100\% & \textbf{80.4\%} & 1 \\
\texttt{K\_online\_learning}   & 3 & Training-as-inference    & \textbf{57\%}   & 100\% & 100\%           & 1 \\
\texttt{L\_diluted}            & 3 & Training-as-inference    & \textbf{52--60\%} & 100\% & \textbf{87.3\%} & 4 \\
\texttt{M\_composite\_memmin}  & 4 & Composite                & 100\%           & 100\% & 100\%           & 4 \\
\texttt{N\_grad\_accum}        & 4 & Training-as-inference    & 100\%           & 100\% & 100\%           & 3 \\
\texttt{O\_composite\_idle\_pad} & 4 & Composite idle-padding & \textbf{19--100\%} & 100\% & 97.6\%       & 11 \\
\texttt{K\_ddp}                & 4 & Multi-GPU DDP            & 100\%           & 100\% & 100\%           & 1 \\
\texttt{K\_ddp\_accum}         & 4 & Multi-GPU DDP            & 99--100\%       & 100\% & 100\%           & 2 \\
\texttt{L\_ddp}                & 4 & Multi-GPU DDP            & \textbf{87--94\%} & 100\% & 99.6\%        & 2 \\
\texttt{L\_ddp\_stagger}       & 4 & Multi-GPU DDP            & \textbf{72\%}   & 100\% & 100\%           & 1 \\
\texttt{whitebox\_full}        & 5 & Whitebox dilution $+$ ckpt & ---          & 100\% & \textbf{87.4\%} & 8 \\
\texttt{whitebox\_lora}        & 5 & Whitebox LoRA $+$ dilution & ---          & 100\% & \textbf{42.6\%} & 3 \\
\bottomrule
\end{tabular}
\end{small}
\end{center}
\vskip -0.1in
\end{table}

\section{Power Validation}
\label{app:power}

While testing 49 workloads on the RTX 5080 desktop workstation, wall-power
readings were collected using a TP-Link Tapo P115 smart plug (readings logged and
timestamped using the \texttt{tapo} Python library).

The P115 smart plug collected workstation power-draw readings at 1\,Hz
simultaneously with NVML. Across 49 runs, we collected 13{,}571 paired,
timestamped readings (median offset 78\,ms, mean offset 128\,ms). The workloads
tested include an idle baseline, 12 non-adversarial workloads (ML training, ML
inference, HPC, cryptomining, rendering) and 36 adversarial runs spanning all
single-GPU strategies implemented in Round 1 through Round 4.

The system idle baseline was 48\,W wall power while the GPU was idle: 26\,W
system overhead $+$ 22\,W for idle GPU based on NVML.

There is a strong correlation between power logged by NVML and wall power, but
the slope of an unfiltered fit is inflated by adversarial strategies that run the
CPU at 180--230\,W (strategies D, K, L, N). Therefore, we report two regressions:
one restricted to the 13 non-adversarial workloads (``clean'', 3{,}698 paired
seconds), and one across all 49 workloads (``all'', 13{,}571 paired seconds):
\begin{align*}
\text{Clean:}\ &\ \text{Wall} \approx 0.97 \times \text{NVML} + 76.1\,\text{W} \\
              &\ (r=0.895,\ R^2=0.80,\ \text{RMSE}=38\,\text{W},\ n=3{,}698) \\[2pt]
\text{All:}\ &\ \text{Wall} \approx 1.50 \times \text{NVML} + 42.7\,\text{W} \\
            &\ (r=0.878,\ R^2=0.77,\ \text{RMSE}=62\,\text{W},\ n=13{,}571)
\end{align*}
The clean dataset's slope of 0.97 confirms that NVML accurately tracks marginal
GPU power, and its intercept of 76\,W reflects the fixed overhead of the
workstation computer while real workloads are running.

Overhead (wall power minus NVML-reported GPU power) averages 107.8 $\pm$ 64.2\,W
for all workloads, or 68--80\,W for non-adversarial ML and HPC workloads,
compared to 180--230\,W for adversarial strategies D, K, L, and N, which trigger
heavy CPU activity while attempting to evade detection through GPU telemetry.
This activity is invisible to NVML but shows up at the wall; the strategies with
heavier CPU overhead are the reason that $R^2$ is 0.77 rather than the 0.80
observed on non-adversarial workloads.

The wall signal is richer than NVML alone, because it captures the host
workstation costs incurred by strategies that manipulate GPU utilization. A
layered monitoring system that combines NVML with wall- or rack-level power
measurements will detect evasion that NVML alone misses.

\section{Roofline Analysis}
\label{app:roofline}

We establish that static architectural heuristics such as the roofline model
\citep{williams2009} are insufficient to detect ML training. We perform a
286-configuration corner-case sweep on H100 NVL, spanning 6 workload families:

\begin{itemize}
\item \textbf{CNN inference} (ResNet-18/50/101/152, 95 configs): enters training
band via batch-size scaling.
\item \textbf{LLM prefill} (GPT-2/M/L, 115 configs): FP32 prefill nearly exactly
matches training arithmetic intensity.
\item \textbf{LLM decode} (GPT-2/M, 30 configs): contrasting case---KV-cache
traffic keeps it memory-bound.
\item \textbf{Quantization variants} (ResNet-50/GPT-2, 18 configs): shifts
arithmetic intensity without changing task semantics.
\item \textbf{Forward-only execution} (ResNet-50, 9 configs): closest semantic
neighbor to training---preserves forward compute while removing
backward/optimizer.
\item \textbf{ViT inference} (ViT-S/B/L, 19 configs): enters training range at
unit batch.
\end{itemize}

Result: 101/286 (35.3\%) non-training configurations fall inside the training
arithmetic-intensity range. Therefore, a temporal telemetry approach is required.

\begin{figure}[h]
\vskip 0.1in
\begin{center}
\centerline{\includegraphics[width=0.92\textwidth]{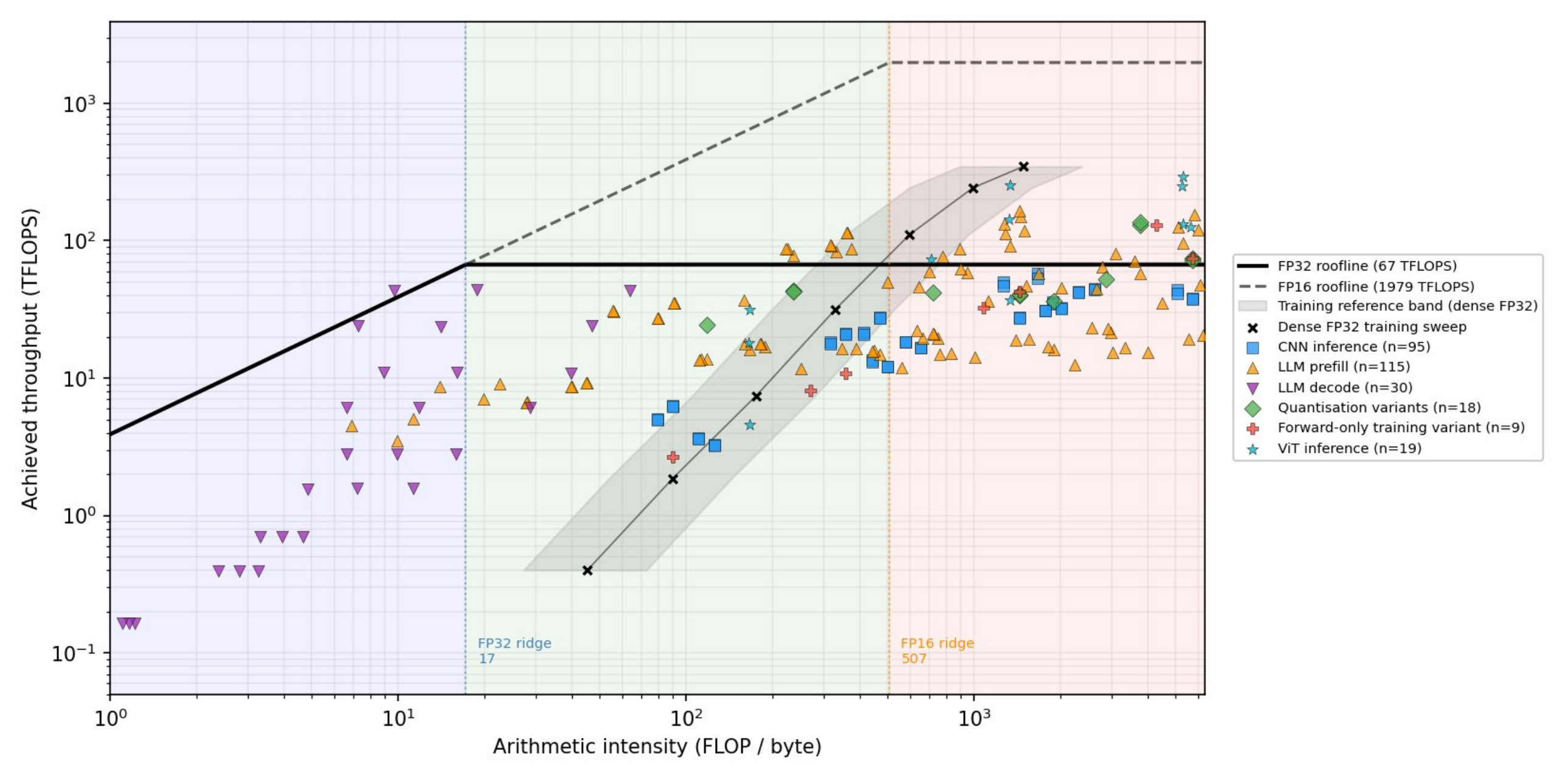}}
\caption{H100-NVL roofline with the 286 corner-case inference configurations
overlaid on a dense-FP32 training reference sweep. Color-coded by workload family
(CNN inference, LLM prefill, LLM decode, quantisation variants, forward-only
training variant, ViT inference). The CNN-inference, LLM-prefill, and
quantisation clouds land squarely inside the training arithmetic-intensity band,
confirming visually that static heuristics cannot separate these inference
workloads from training.}
\label{fig:roofline}
\end{center}
\vskip -0.1in
\end{figure}

\section{Window Size vs.\ Accuracy}
\label{app:window}

Classification accuracy is compared across 5/10/15/20/30/45/60/90/120\,s windows.
Detection is essentially flat from 15\,s onward (98.2--99.0\%); 5\,s is 98.15\%
and 10\,s is 98.81\%, so lowering the detection window to 5\,s costs less than 1
percentage point. A monitoring system can therefore operate with $\leq$15\,s
alert latency without meaningful accuracy loss.

\begin{figure}[h]
\vskip 0.1in
\begin{center}
\centerline{\includegraphics[width=0.55\textwidth]{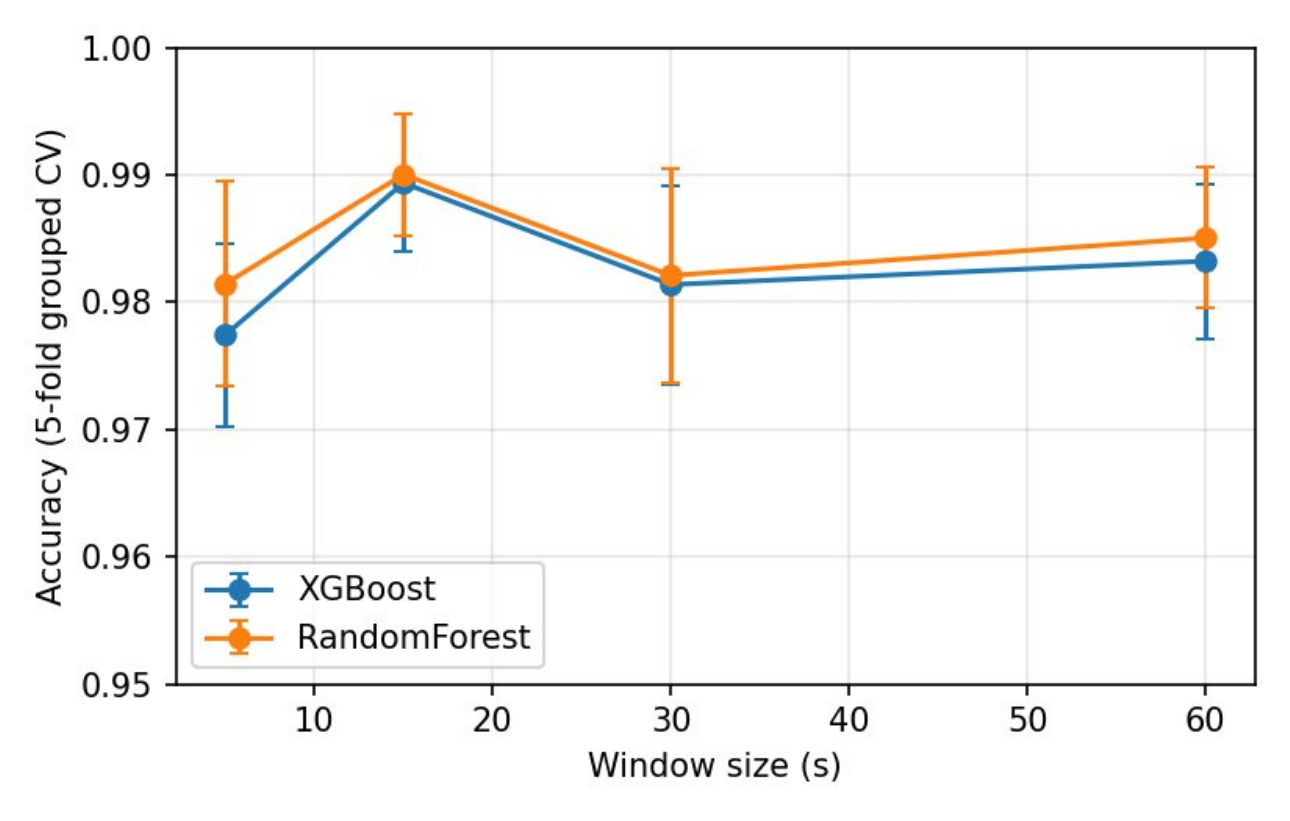}}
\caption{Accuracy vs.\ window size.}
\label{fig:window}
\end{center}
\vskip -0.1in
\end{figure}

\section{Additional Results}
\label{app:additional}

\begin{figure}[h]
\vskip 0.1in
\begin{center}
\centerline{\includegraphics[width=0.92\textwidth]{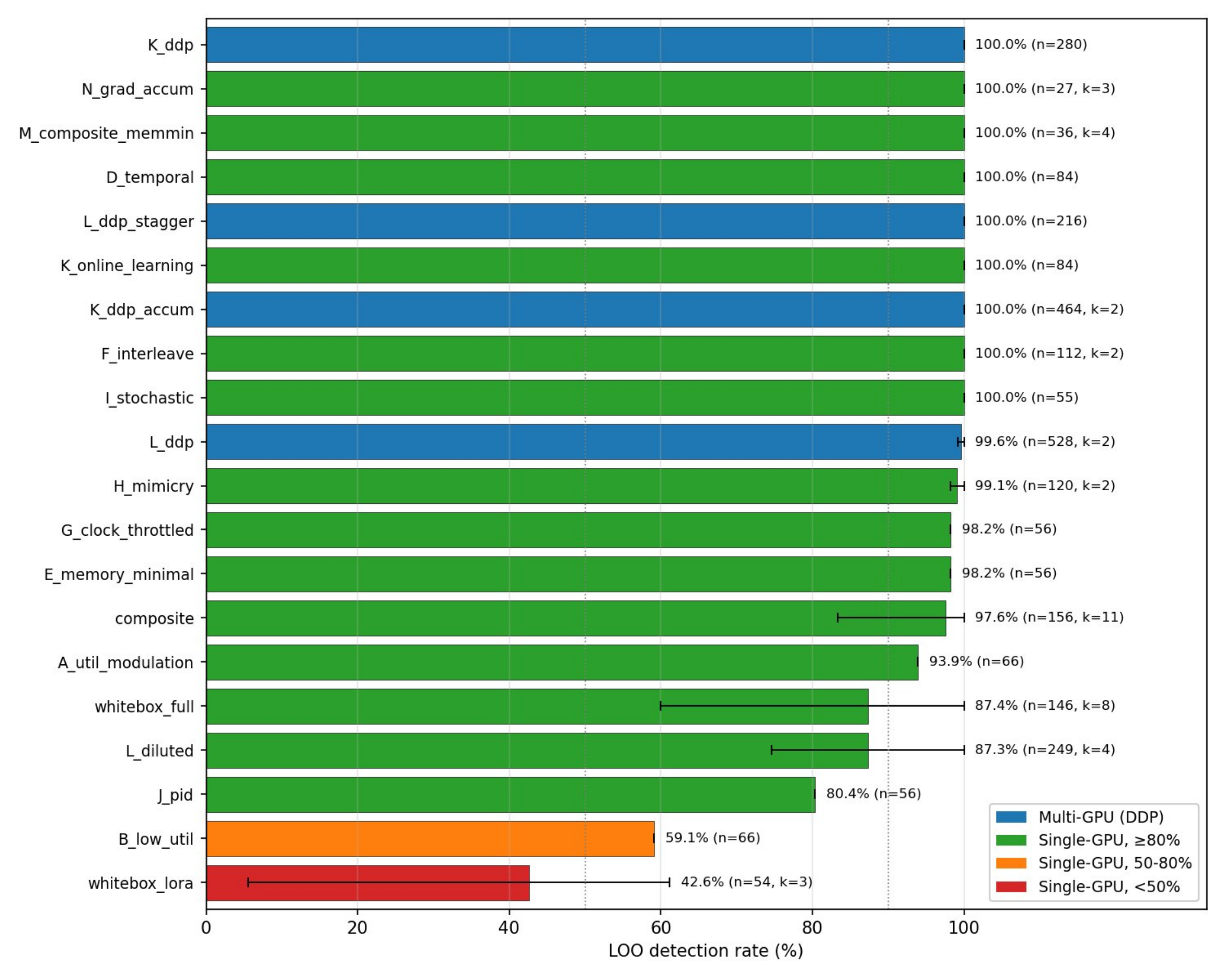}}
\caption{Detection rate in LOO evaluation by strategy family. Bar chart showing
20 families sorted by detection rate, with error bars showing min/max across
variants within each family.}
\label{fig:lood}
\end{center}
\vskip -0.1in
\end{figure}

\section{Limitations and Future Work}
\label{app:limitations}

This work has several limitations. First, our experiments were conducted
exclusively on single-node hardware configurations, testing workloads on
instances up to only 8 GPUs. Future experiments should expand to greater GPU
counts, including multi-node clusters. Evasion characteristics for multi-node
model- and data-parallel training will differ due to different types of
interconnect (such as InfiniBand between servers vs.\ NVLink within servers), but
it is likely that training will continue to be highly distinctive due to the
communication requirements of large-scale training. Relatedly, model sizes were
limited by memory constraints, with largest usable models of 70B parameters for
training and 671B MoE for inference. Today's frontier models are often much
larger and can require more than 8 GPUs to train and serve.

Second, our adversarial strategies were limited to framework-level evasion
implemented in PyTorch. An adversary with access to custom CUDA kernels could
potentially manipulate individual kernel behavior, which we did not test. Future
research should investigate the effectiveness of these strategies and see how
robust the classifier becomes against them.

Third, the integrity of the telemetry channel is a critical assumption. Our
evaluation assumes the adversary cannot tamper with NVML readings or performance
counter values, nor inject synthetic telemetry to spoof a workload signature.
These guarantees require a HEM for tamper-resistant telemetry and an
authenticated channel. Other work like \citet{monfared2026} investigates what
can be done if these guarantees are not available.

Fourth, we only tested on NVIDIA GPUs released since 2020. This category includes
the large majority of compute capacity available to cloud tenants, but
substantial amounts of other chips exist, including AMD and Huawei GPUs and
custom silicon such as Google TPUs and Amazon Inferentia and Trainium
\citep{epoch2025chips}. Future work should generalize this approach using the
analogous telemetry signals available for other hardware.

Fifth, our telemetry was generally sampled at $\sim$1\,Hz, and features were
extracted from a limited variety of window sizes. Future work should test other
sampling rates, measuring how much classification improves with more frequent
sampling, how much sampling can be done without degrading performance, and
investigating the minimum sample rate that is still difficult to evade.

Finally, we only tested a limited range of workloads and adversarial strategies.
Real datacenters run a wider variety of applications than we emulated, with
notable gaps including multi-user inference serving, data processing and ETL
pipelines, and predictive analytics. Although we tried to cover many evasion
strategies and edge cases, real evaders might attempt a much wider range of
tactics that we cannot anticipate. It is important to continue iterating on this
work to make the classifier even more robust to unknown future threats.
Iteration should include more white-box rounds, to determine whether resistance
to white-box attacks improves generally, or if an evader can always defeat a
classifier if they obtain white-box access, and if so, how much performance
penalty is incurred in the process.

\end{document}